\documentclass[letterpaper, 10 pt, conference]{ieeeconf}  

\IEEEoverridecommandlockouts                              

\usepackage{amsmath} 
\usepackage{amssymb}  
\usepackage{threeparttable}
\usepackage{booktabs}
\usepackage{multirow}
\usepackage{bbding}
\UseRawInputEncoding

\usepackage{graphicx}
\usepackage{subfigure}
\graphicspath{{pdfs/},{images/} }

\newcommand{\figref}[1]{Fig.~\ref{#1}}
\newcommand{\tabref}[1]{Table~\ref{#1}}
\newcommand{\secref}[1]{Section~\ref{#1}}

\title{\LARGE \bf
SP-VINS: A Hybrid Stereo Visual Inertial Navigation System based on Implicit Environmental Map
}

\author{Xueyu Du, Lilian Zhang$^{*}$, Fuan Duan, Xincan Luo, Maosong Wang, Wenqi Wu, and JunMao
	\thanks{Authors are with the College of Intelligent Science and Technology, National University
		of Defense Technology, Changsha, 410073, China.}%
	\thanks{$^{*}$ Correspondence to: lilianzhang@nudt.edu.cn.}%
	\thanks{This research is funded by the National Natural Science Foundation of China (grant number: 62103430, 62103427, 62073331) and Major Project of Natural Science Foundation of Hunan Province (No. 2021JC0004).}
}

\begin{document}

\maketitle
\thispagestyle{empty}
\pagestyle{empty}

\begin{abstract}
Filter-based visual inertial navigation system (VINS) has attracted mobile-robot researchers for the good balance between accuracy and efficiency, but its limited mapping quality hampers long-term high-accuracy state estimation.
To this end, we first propose a novel filter-based stereo VINS, differing from traditional simultaneous localization and mapping (SLAM) systems based on 3D map, which performs efficient loop closure constraints with implicit environmental map composed of keyframes and 2D keypoints.
Secondly, we proposed a hybrid residual filter framework that combines landmark reprojection and ray constraints to construct a unified Jacobian matrix for measurement updates.
Finally, considering the degraded environment, we incorporated the camera-IMU extrinsic parameters into visual description to achieve online calibration.
Benchmark experiments demonstrate that the proposed SP-VINS achieves high computational efficiency while maintaining long-term high-accuracy localization performance, and is superior to existing state-of-the-art (SOTA) methods.  

\end{abstract}

\section{INTRODUCTION}
Visual-inertial (VI) configuration has become increasingly prominent in autonomous robot navigation, because complimentary sensing nature and good balance between cost and accuracy.
When the signal of Global Navigation Satellite System (GNSS) is weak or denied, VINS can provide accurate state estimation for mobile-robot \cite{WOS:000494942307003,WOS:000446394502008,cadena2017past}. 

As shown in \cite{cadena2017past} and \cite{WOS:000725804900006}, VINS can adopt the following three types of data association patterns to obtain real-time state estimation or construct environmental map. 

The first type is short-term data association, with representative works are MSCKF-based visual inertial odometry (VIO) \cite{4209642,WOS:000424646100016}.
This method does not construct map points for long-term observation, which achieves high computational efficiency by regularly marginalizing used or tracking lost visual measurements and related poses. 
However, the drawback is that even if the system moves in the same area, accumulated estimation drift cannot be suppressed \cite{WOS:000725804900006}.

The second type is mid-term data association, which involves jointly estimating local map points and poses. 
Representative works include hybrid MSCKF (combined MSCKF and EKF-SLAM) \cite{9196524,fan2024schurvins,yuan2025voxel} and optimization-based VIO \cite{WOS:000350472800005,8594007}.
This method can utilize the local environmental structure to achieve better state estimation, but it still lacks global localization performance in large-scale scenarios \cite{WOS:000725804900006}.

The third type is long-term data association. When the system revisits a certain area, historical poses and map estimation are used to correct accumulated navigation drift, this process known as loop closure (LC) \cite{tsintotas2022revisiting} and can help robots achieve long-term high-accuracy localization and mapping.
Representative works include ORB-SLAM3 \cite{WOS:000725804900006}, Kimera \cite{WOS:000712319501044}, VINS-Fusion \cite{qin2019general}, etc.
However, it requires continuous data association between the current observed and retained historical map and keyframe poses, which will consume a large amount of computing resources in long-term and large-scale tasks \cite{schmidt2025visual,geneva2022map}.

Above all, each of the three data association patterns has its own advantages in terms of accuracy and efficiency.
Therefore, integrating and improving based on the three patterns will help VINS provide long-term, high-accuracy state estimation while maintaining low computational resource consumption.

In this work, we propose a combination of short-term and long-term data associations:
We first combine the double state transformation extended Kalman filter (DST-EKF), pose-only (PO) visual description, multi-state constrained ray estimation and online extrinsic calibration for constructing a stereo VIO framework to provide consistent and efficient state estimation in both short-term or open-loop motions;
Secondly, in response to the high computational consumption problem of traditional 3D map based SLAM systems, a improved long-term data association strategy based on implicit map is introduced to achieve more efficiently global drift correction.
In conclusion, the contributions of this letter can be summarized as follows:

\begin{figure*}[htbp]
	\centering
	\subfigure[ ]{\centering \includegraphics [width=.90\columnwidth]{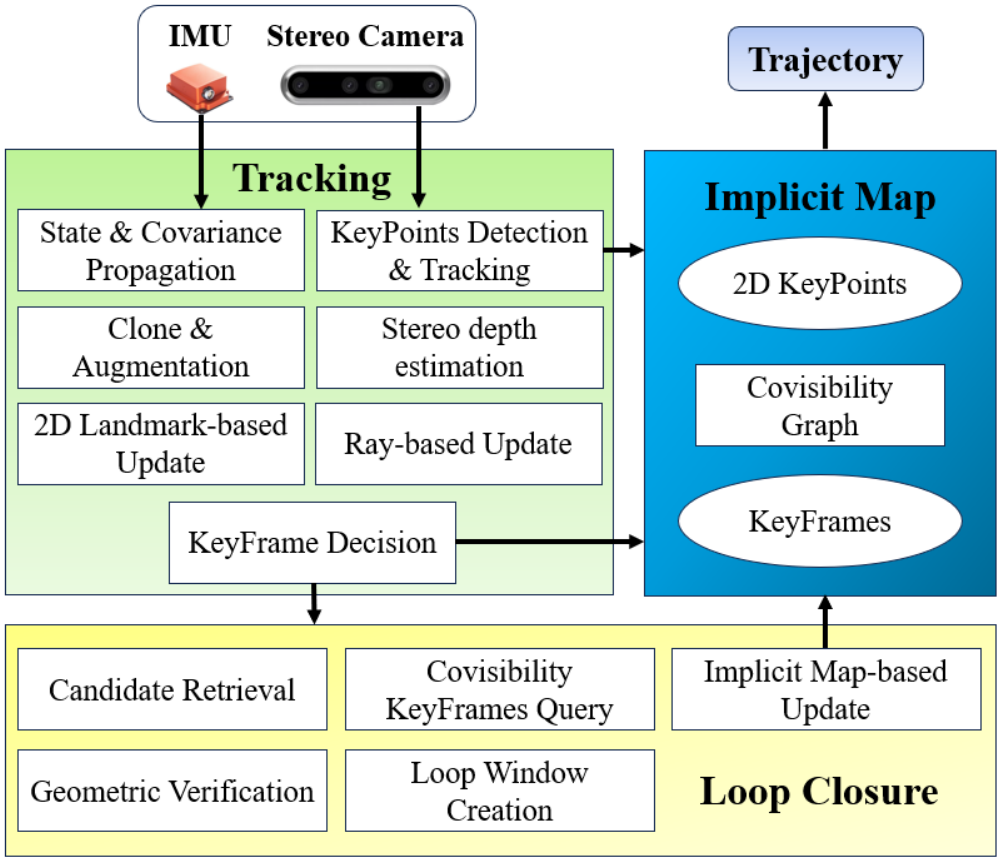}\label{fig:overview1}}
	\subfigure[ ]{\centering \includegraphics [width=.90\columnwidth]{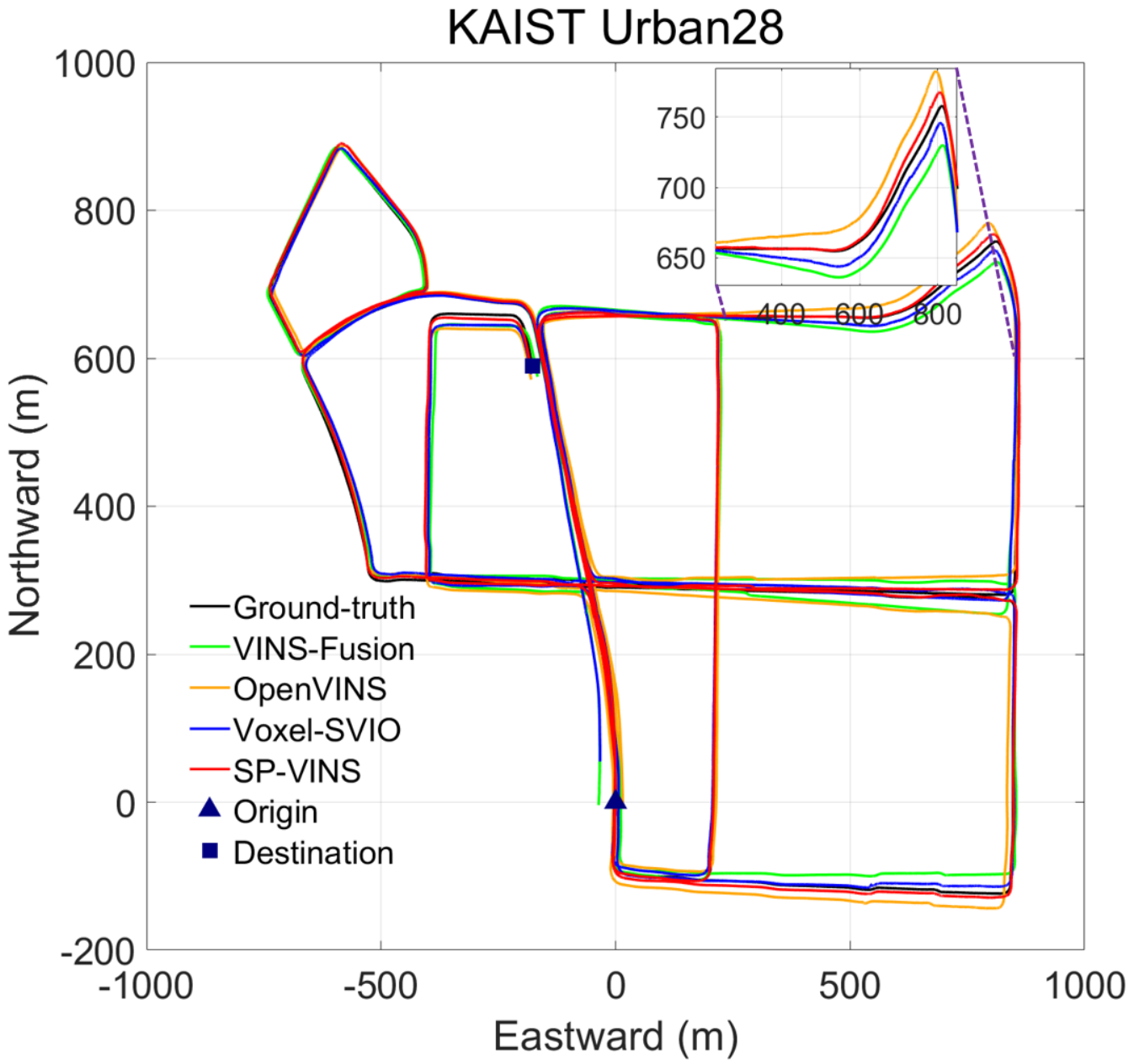}\label{fig:overview2}}
	\caption{
		(a) Framework of SP-VINS, which differs from traditional 3D map based SLAM systems, only performs global drift correction based on implicit environmental map composed of keyframes and 2D keypoints;
		(b) Comparison of SP-VINS with three SOTA methods on sequence Urban28 \cite{jeong2019complex}.
	}
	\label{fig:overview}
\end{figure*}

\begin{itemize}
	\item We propose a lightweight relocalization framework that adopts the implicit environmental map composed of keyframes and 2D keypoints instead of 3D map to achieve long-term loop closure and recovery from navigation drift.
	\item We propose a consistent and efficient hybrid VIO framework, which takes the DST-EKF based inertial model as the skeleton and combines the pose-only landmark reprojection and multi-state constrained ray estimation to construct a hybrid residual model. 
	\item We incorporate the installation relationship of camera-IMU into the PO visual description to achieve online calibration in the degraded environment.
	\item We propose a novel filter-based stereo VINS based on implicit map, which does not rely on 3D map and pose graph optimization (PGO) to achieve long-term high-accuracy state estimation. The experimental results from different public datasets show that SP-VINS significantly improves the accuracy and efficiency compared with the SOTA method.
\end{itemize}

\section{Related Work}
Visual-inertial navigation system can be classified into VIO and VI-SLAM, depending on whether it estimates ego-motion while constructing and maintaining a globally consistent environmental map.  

\subsection{Visual Inertial Odometry}
The efforts of various researchers have led to the emergence of many excellent VIO systems. 
MSCKF \cite{4209642} is a classic filter-based VIO framework that adopts multiple historical poses and related visual measurements to constrain and estimate ego-motion. Benefiting from the low cost and high efficiency, MSCKF has been widely deployed and derived many excellent works \cite{WOS:000424646100016,9196524,fan2024schurvins,yuan2025voxel}.
OpenVINS \cite{9196524} suppress the drift accumulated over time by combining MSCKF and EKF-SLAM, which selects appropriate 3D features for joint state estimation.
Furthermore, to ensure rational allocation of computing resources, \cite{yuan2025voxel} introduced voxel map management, and \cite{yuan2025voxel} adopted Schur complement, enabling the system to manage and apply 3D map points more efficiently.
Additionally, as a representative work of optimization-based VIO, \cite{WOS:000350472800005} integrates tightly coupled IMU and visual measurements into the factor graph to achieve accurate pose estimation, while controlling computational resources consumption through marginalization and keyframe mechanisms.
Despite this, since the VIO system does not build and maintain a global environmental map, it cannot effectively utilize historical information to correct the drift accumulated over time in large-scale scenarios.
\subsection{Visual Inertial SLAM}
Compared with VIO, VI-SLAM performs loop closure detection during the local pose estimation process. After identifying the previously visited areas, it can utilize the constructed environmental map to achieve global drift correction.
VINS-Fusion \cite{qin2019general} adopts the bag-of-words (DBoW2) based on BRIEF descriptors to identify previously visited locations, and then employs PGO to reduce drift over time.
ORB-SLAM3 \cite{WOS:000725804900006} first adopts the DBoW2 based ORB descriptors to detect loop, then employs PGO to correct drifts quickly, and finally starts the global bundle adjustment (GBA) to optimize all keyframes and map points.
Typical VI-SLAM works also include \cite{WOS:000712319501044,xu2025airslam,usenko2019visual}, etc.

Most popular VI-SLAM systems are optimization-based methods, which achieve high-accuracy localization performance but also require substantial computing resources \cite{WOS:000446394502008}.
Filter-based methods offer high efficiency but are limited by relatively low mapping quality, which affects long-term navigation performance during location revisits \cite{zhao2020closed}.
Above all, if filter-based method can be decoupled from inaccurate 3D map point estimation, it will be helpful to achieve a VI-SLAM work that strikes a good balance between accuracy and efficiency.
\cite{WOS:000458768000003,WOS:000899419900005} proposes a novel PO visual description, which allows for the equivalent 3D feature representation of using only the camera pose and corresponding 2D measurements.  
\cite{du2025spvio,wang2025po} adopted this theory to reconstruct the MSCKF framework, decoupling system from 3D map points while obtaining more accurate and efficient state estimation.
However, all these methods are VIO systems that do not account for long-term data association.

\section{Overview}
\begin{figure}[htbp]
	\centering
	\includegraphics [width=3.4in]{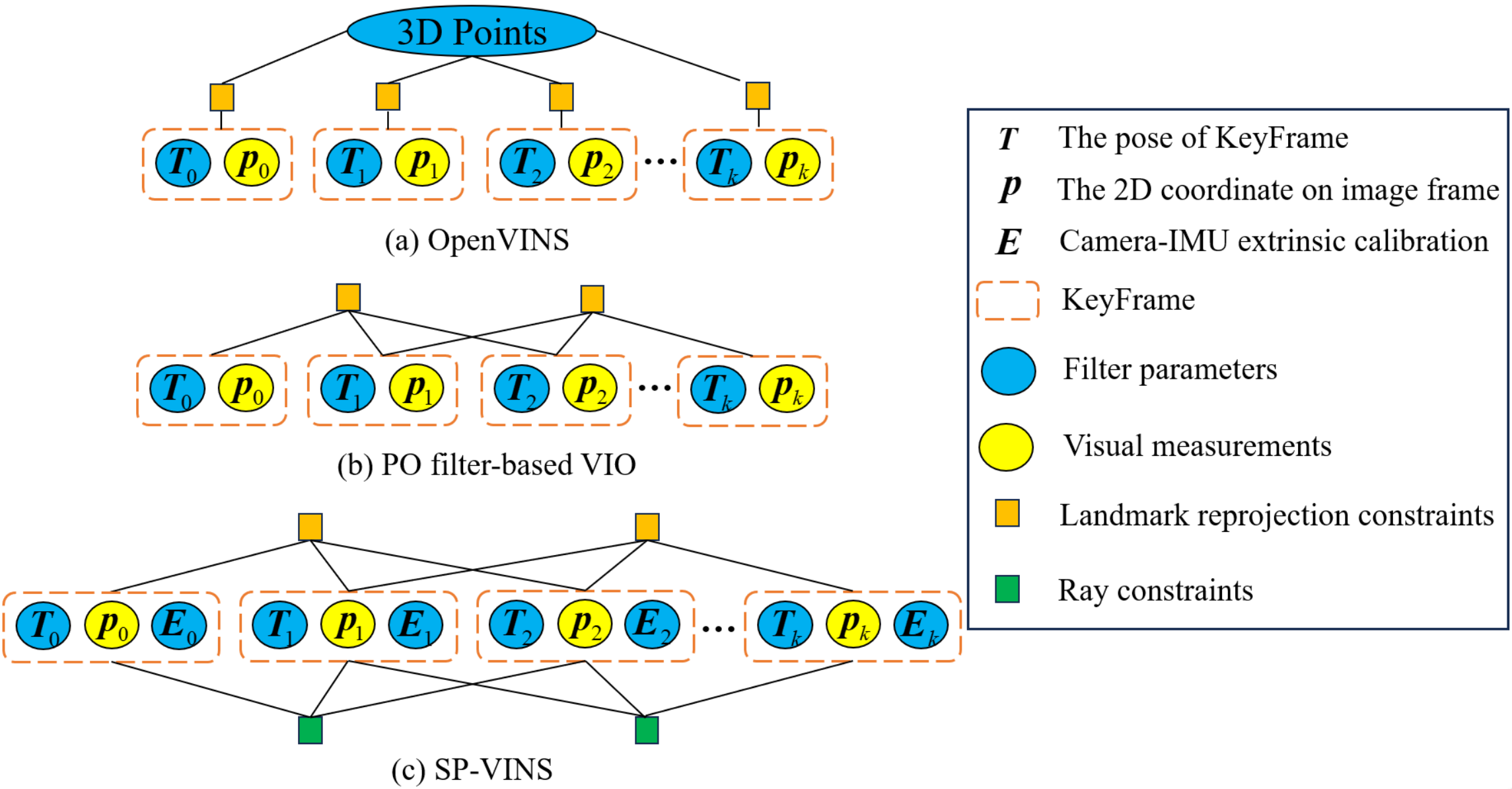} 
	\caption{
		Comparison among the visual residual representation of OpenVINS \cite{9196524}, PO filter-based VIO \cite{du2025spvio,wang2025po} and SP-VINS.
	}
	\label{fig:residual representation}
\end{figure}

The structure of proposed stereo visual inertial navigation system is shown in \figref{fig:overview1}, with its primary components outlined in the subsequent subsections.
The tracking thread adopts tightly coupled sensor measurements to perform local pose estimation, and simultaneously selects keyframes and corresponding 2D keypoints to update the implicit environment map.
As shown in \figref{fig:residual representation}, different from OpenVINS \cite{9196524} and PO filter-based VIO \cite{du2025spvio,wang2025po}, SP-VINS adopts a hybrid residual representation combining landmark reprojection and ray constraints for local pose estimation, and introduces the online camera-IMU extrinsic calibration function.
The loop closure thread detects loops and utilizes implicit map to reduce cumulative errors.
Unlike 3D map-based SLAM systems \cite{qin2019general}, SP-VINS utilizes the covisibility relationship between active states and loops to update estimated state instead of performing PGO and GBA.

We now define notations and frame definitions that we use throughout this letter.
We consider $\left\{G\right\}$ as the global frame. The direction of gravity is aligned with $z$-axis of the global frame, and $^{G}\boldsymbol{g}=\begin{bmatrix}
	0 & 0 & g
\end{bmatrix}^{T}$ represents the gravity vector in the global frame.
$\left\{B\right\}$ is the body frame, which we define to be the same as IMU frame.
$\left\{C\right\}$ is the camera frame, while $\left\{C_L\right\}$ and $\left\{C_R\right\}$ respectively represent the left and right camera frame in stereo visual system.
We adopt both rotation matrices $\boldsymbol{R}$ and Hamilton quaternions $\boldsymbol{q}$ to represent rotation, and then adopt $\boldsymbol{\phi}$ to define corresponding error angle.
$^{G}_{B}\boldsymbol{q}$ and $^{G}\boldsymbol{p}_{B}$ are the rotation and translation from body frame to world frame (The pose transformation between other frames is similar).
$\left\{B_k\right\}$ and $\left\{C_k\right\}$ are the body frame and camera frame at time $k$.
$\otimes$ represents the quaternion multiplication, and $\left[\cdot \times\right]$ is the antisymmetric matrix.
$\boldsymbol{e}_1$, $\boldsymbol{e}_2$ and $\boldsymbol{e}_3$ represent the standard basis vectors in 3D space respectively.
$\delta \left[\cdot\right]$ and $\dot{[\cdot]}$ respectively represent the error or estimation of a certain quantity.
\section{SP-VINS Framework}
\subsection{State definitions}
The state vector of our visual-inertial system at time $k$ can be defined as $\boldsymbol{x}_{k}$, which consists of the current active state $\boldsymbol{x}_{A}$ and a set of $m$ map keyframe poses $\boldsymbol{x}_{K}$. The active state $\boldsymbol{x}_{A}$ contain the current inertial navigation state $\boldsymbol{x}_{b}$, $n$ historical IMU pose clones $\boldsymbol{x}_{c}$ and camera-IMU extrinsic calibration $\boldsymbol{x}_{e}$.
\begin{align}
	\boldsymbol{x}_{k}&=\begin{bmatrix} \boldsymbol{x}_{A}^{T} & \boldsymbol{x}_{K}^{T} \end{bmatrix} ^{T} = \begin{bmatrix}
		\boldsymbol{x}_{b}^{T} & \boldsymbol{x}_{c}^{T} & \boldsymbol{x}_{e}^{T} & \boldsymbol{x}_{K}^{T}
	\end{bmatrix}^{T} \label{eqn:x state}\\
	\boldsymbol{x}_{b}&=\begin{bmatrix} _{{B}_{k}}^{G}\boldsymbol{q}^{T} & ^{G}\boldsymbol{v}_{{B}_{k}}^{T} & ^{G}\boldsymbol{p}_{{B}_{k}}^{T} & \boldsymbol{b}_{{g}_{k}}^{T} & \boldsymbol{b}_{{a}_{k}}^{T} \end{bmatrix} ^{T} \label{eqn:xb state}\\
	\boldsymbol{x}_{c}&=\begin{bmatrix} _{{B}_{k-n}}^{G}\boldsymbol{q}^{T} & ^{G}\boldsymbol{p}_{{B}_{k-n}}^{T} & {...} & _{{B}_{k-1}}^{G}\boldsymbol{q}^{T} & ^{G}\boldsymbol{p}_{{B}_{k-1}}^{T} \end{bmatrix} ^{T} \label{eqn:xc state} \\
	\boldsymbol{x}_{e}&=\begin{bmatrix} _{B}^{C_L}\boldsymbol{q}^{T} & ^{C_L}\boldsymbol{p}_{B}^{T} & _{B}^{C_R}\boldsymbol{q}^{T} & ^{C_R}\boldsymbol{p}_{B}^{T} \end{bmatrix} ^{T} \label{eqn:xe state} \\
	\boldsymbol{x}_{K}&=\begin{bmatrix} _{{B}_{1}}^{G}\boldsymbol{q}^{T} & ^{G}\boldsymbol{p}_{{B}_{1}}^{T} & {...} & _{{B}_{m}}^{G}\boldsymbol{q}^{T} & ^{G}\boldsymbol{p}_{{B}_{m}}^{T} \end{bmatrix} ^{T} \label{eqn:xK state}
\end{align}
where $\boldsymbol{b}_{{g}_{k}}$ and $\boldsymbol{b}_{{a}_{k}}$ are the gyroscope and accelerometer biases, and $^{G}\boldsymbol{v}_{{b}_{k}}$ is the body frame velocity in global frame.

From Eq. \eqref{eqn:x state} to Eq. \eqref{eqn:xK state}, the corresponding error-state of $\boldsymbol{x}_{k}$ can be defined as:
\begin{align}
	\delta \boldsymbol{x}_{k}&= \begin{bmatrix}
		\delta \boldsymbol{x}_{A}^{T} & \delta \boldsymbol{x}_{K}^{T}
	\end{bmatrix}^{T} =\begin{bmatrix} \delta \boldsymbol{x}_{b}^{T} & \delta \boldsymbol{x}_{c}^{T} & \delta \boldsymbol{x}_{e}^{T} & \delta \boldsymbol{x}_{K}^{T} \end{bmatrix} ^{T} \label{eqn:delta x state}
\end{align}

The extended additive error of quaternion is defined as follow:
\begin{equation}\label{eqn:quaternions}
	\begin{aligned}
		\boldsymbol{q}&=\delta \boldsymbol{q} \otimes \boldsymbol{\widetilde{q}} \\
		{\delta \boldsymbol{q}}&= \begin{bmatrix}
			1 & \frac{1}{2} \boldsymbol{\phi}^{T} 
		\end{bmatrix}^{T}	
	\end{aligned}
\end{equation}
where the extended additive error of rotation matrix can be expressed as:
\begin{equation}\label{eqn:rotation error}
	\begin{aligned}
		\boldsymbol{R}(\boldsymbol{q}) &= \boldsymbol{R} \\  
		\boldsymbol{\widetilde{R}} &= (\boldsymbol{I}-[\boldsymbol{\phi} \times])\boldsymbol{R}
	\end{aligned}
\end{equation}

Except for velocity and position, other states can be used with the error definition (e.g. $\boldsymbol{x} = \boldsymbol{\widetilde{x}} + \delta \boldsymbol{x}$). 
According to \cite{du2025spvio}, after the state transformation based on Lie-group theory, the new velocity and position errors can be defined as:
\begin{align}\label{eqn:error_state_DST}
	\left\{
	\begin{aligned}
		{\delta ^{G}\boldsymbol{v}^{(ST)}_{B}} &= -{\delta {^{G}\boldsymbol{v}_{B}}} + [{^{G}\boldsymbol{v}_{B}} \times]{_{B}^{G}\boldsymbol{\phi}} \\
		{\delta ^{G}\boldsymbol{p}_{B}^{(ST)}} &= -{\delta {^{G}\boldsymbol{p}_{B}}} + [{^{G}\boldsymbol{p}_{B}} \times]{_{B}^{G}\boldsymbol{\phi}} 
	\end{aligned}
	\right.
\end{align}

\subsection{Propagation and Augmentation}
SP-VINS follows the policy introduced in \cite{du2025spvio} to perform propagation and augmentation for the current active state $\boldsymbol{x}_{A}$.
The linearized continuous dynamics for the error IMU state is defined as:
\begin{align}\label{eqn:deltaxb process PO}
	\delta \boldsymbol{\dot{x}}_{b}&=\boldsymbol{F}_{b} \delta \boldsymbol{x}_{b}+\boldsymbol{G}_{b}\boldsymbol{w}_{b} 
\end{align}
where $\boldsymbol{w}_{b}=\begin{bmatrix} \boldsymbol{w}_{g} & \boldsymbol{w}_{a} & \boldsymbol{w}_{wg} & \boldsymbol{w}_{wa} \end{bmatrix} ^{T}$ is the system noise vector.  
$\boldsymbol{w}_{g}$ and $\boldsymbol{w}_{a}$ represent the measurement white noise of the gyro and accelerometer respectively; $\boldsymbol{w}_{wg}$ and $\boldsymbol{w}_{wa}$ represent the driven white noise of the gyro biases and accelerometer biases.
 
The error-state transition matrix $\boldsymbol{F}_{b}$ and the input noise Jacobian matrix $\boldsymbol{G}_{b}$ can be represented as:
\begin{footnotesize}
	\begin{align}\label{eqn:F PO}
		\boldsymbol{F}_{b}=
		\begin{bmatrix}
			- [\boldsymbol{\omega}^{G}\times] & \boldsymbol{0}_{3} & \boldsymbol{0}_{3} & -_{B}^{G}\boldsymbol{R} & \boldsymbol{0}_{3} \\
			[\boldsymbol{g}^{G}\times]\\+[^{G}\boldsymbol{v}_{B}\times][\boldsymbol{\omega}^{G}\times] & -2 [\boldsymbol{\omega}^{G}\times] & \boldsymbol{0}_{3} & -[^{G}\boldsymbol{v}_{B}\times]{_{B}^{G}\boldsymbol{R}} & -_{B}^{G}\boldsymbol{R} \\
			-[^{G}\boldsymbol{p}_{B}\times][\boldsymbol{\omega}^{G}\times] & \boldsymbol{I}_{3} & \boldsymbol{0}_{3} & -[^{G}\boldsymbol{p}_{B}\times]{_{B}^{G}\boldsymbol{R}} &\boldsymbol{0}_{3} \\
			\boldsymbol{0}_{3} & \boldsymbol{0}_{3} & \boldsymbol{0}_{3} & \boldsymbol{0}_{3} &\boldsymbol{0}_{3} \\
			\boldsymbol{0}_{3} & \boldsymbol{0}_{3} & \boldsymbol{0}_{3} & \boldsymbol{0}_{3} &\boldsymbol{0}_{3}
		\end{bmatrix}
	\end{align}
\end{footnotesize} 
\begin{align}\label{eqn:G PO}
	\boldsymbol{G}_{b}=
	\begin{bmatrix}
		-_{B}^{G}\boldsymbol{R} & \boldsymbol{0}_{3} & \boldsymbol{0}_{3} & \boldsymbol{0}_{3} \\
		-[^{G}\boldsymbol{v}_{B}\times]{_{B}^{G}\boldsymbol{R}} & -{_{B}^{G}\boldsymbol{R}} & \boldsymbol{0}_{3} & \boldsymbol{0}_{3} \\
		-[^{G}\boldsymbol{p}_{B}\times]{_{B}^{G}\boldsymbol{R}} & \boldsymbol{0}_{3} & \boldsymbol{0}_{3} & \boldsymbol{0}_{3} \\
		\boldsymbol{0}_{3} & \boldsymbol{0}_{3} & \boldsymbol{I}_{3} & \boldsymbol{0}_{3} \\
		\boldsymbol{0}_{3} & \boldsymbol{0}_{3} & \boldsymbol{0}_{3} & \boldsymbol{I}_{3} 
	\end{bmatrix}
\end{align} 

Define the state covariance matrix at time $k$ as $\boldsymbol{P}_{k}$, which can be expressed by:
\begin{align}\label{eqn:Pk}
	\boldsymbol{P}_{k} =
	\begin{bmatrix}
		\boldsymbol{P}_{II} & \boldsymbol{P}_{IO} \\
		\boldsymbol{P}_{IO}^{T} & \boldsymbol{P}_{OO}
	\end{bmatrix}
\end{align}
where $\boldsymbol{P}_{IO}$ is the covariance of inertial state, $\boldsymbol{P}_{IO}$ is the covariance between inertial state other states, $\boldsymbol{P}_{OO}$ is covariance of other states.

After discretizing the continuous system to obtain discrete transition matrix $\boldsymbol{\Phi}$ and noise-driven matrix $\boldsymbol{Q}$, the covariance propagation process can be expressed as:
\begin{equation}\label{eqn:cov propagation}
	\begin{aligned}
		\boldsymbol{P}_{II} &= {\boldsymbol{\Phi}}\boldsymbol{P}_{II}{\boldsymbol{\Phi}^{T}} + \boldsymbol{Q} \\
		\boldsymbol{P}_{IO} &= {\boldsymbol{\Phi}}{\boldsymbol{P}_{IO}}
	\end{aligned}
\end{equation}

The state and covariance augmentation are similar with OpenVINS \cite{9196524} and will not be repeated here.
Additionally, since the application of Lie-group-based state transformation to reconstruct velocity and position states, the relevant state update process is modified as:
\begin{small}
	\begin{align}\label{eqn:delta update}
		&\begin{bmatrix}
			{_{B}^{G}\boldsymbol{R}} & {^{G}\boldsymbol{v}_{B}} & {^{G}\boldsymbol{p}_{B}} \\
			\boldsymbol{0}_{1\times3} & 1 & 0 \\
			\boldsymbol{0}_{1\times3} & 0 & 1 \\
		\end{bmatrix} = \\ \nonumber 
		&\begin{bmatrix}
			[\boldsymbol{I}+{_{B}^{G}\boldsymbol{\phi}}\times ] & \delta{^{G}\boldsymbol{v}_{B}} & \delta{^{G}\boldsymbol{p}_{B}} \\
			\boldsymbol{0}_{1\times3} & -1 & 0 \\
			\boldsymbol{0}_{1\times3} & 0 & -1 \\
		\end{bmatrix}
		\begin{bmatrix}
			{_{B}^{G}\boldsymbol{\widetilde{R}}} & {^{G}\boldsymbol{\widetilde{v}}_{B}} & {^{G}\boldsymbol{\widetilde{p}}_{B}} \\
			\boldsymbol{0}_{1\times3} & -1 & 0 \\
			\boldsymbol{0}_{1\times3} & 0 & -1 \\
		\end{bmatrix}
	\end{align}
\end{small}

\subsection{2D Landmark-based Visual Residual}
SP-VINS adopt the PO theory \cite{WOS:000899419900005} to construct 2D landmark-based visual residual, which serves as the theoretical foundation for VINS that decouple from 3D map.

Assuming a 3D landmark $^{G}\boldsymbol{p}_{f}=\begin{bmatrix} ^{G}{X}_{f} & ^{G}{Y}_{f} & ^{G}{Z}_{f} \end{bmatrix}^{T}$ observed in $n$ images, its normalized coordinate in the $i$-th image is $\boldsymbol{p}_{{C}_{i}}=\begin{bmatrix} {u}_{i} & {v}_{i} & 1 \end{bmatrix}^{T}$ ($i=1,...n$).
According to the PO visual description \cite{WOS:000899419900005}, the projection of 3D landmark $^{G}\boldsymbol{p}_{f}$ in $i$-th image can be expressed as:
\begin{small}
	\begin{equation}
		\begin{aligned}\label{eqn:3Dpos PO}
			^{{C}_{i}}\boldsymbol{p}_{{f}}&= {_{G}^{{C}_{i}}\boldsymbol{R}}({^{G}\boldsymbol{p}_{f}}-{^{G}\boldsymbol{p}_{{C}_{i}}}) \\
			&= ||[{^{{C}_{\beta}}\boldsymbol{t}_{{C}_{\alpha}}}\times]\boldsymbol{p}_{{C}_{\beta}}||{_{{C}_{\alpha}}^{{C}_{i}}\boldsymbol{R}}\boldsymbol{p}_{{C}_{\alpha}}+||[{\boldsymbol{p}_{{C}_{\beta}}}\times]{_{{C}_{\alpha}}^{{C}_{\beta}}\boldsymbol{R}}{\boldsymbol{p}_{{C}_{\alpha}}}||{^{{C}_{i}}\boldsymbol{t}_{{C}_{\alpha}}}
		\end{aligned}
	\end{equation}
\end{small}
where ${_{{C}_{\alpha}}^{{C}_{\beta}}\boldsymbol{R}}$ and ${^{{C}_{\beta}}\boldsymbol{t}_{{C}_{\alpha}}}$ represent the pose transformation relationship of two camera frames. Combined with the camera-IMU extrinsic calibration, they can be specially expressed as:
\begin{equation}\label{eqn:R CkCj}
	\begin{aligned}
		{_{{C}_{\alpha}}^{{C}_{\beta}}\boldsymbol{R}} ={^{C_\beta}_{G}\boldsymbol{R}}{_{C_\alpha}^{G}\boldsymbol{R}}={^{C}_{B}\boldsymbol{R}}{^{B_\beta}_{G}\boldsymbol{R}}{_{B_\alpha}^{G}\boldsymbol{R}}{_{C}^{B}\boldsymbol{R}}
	\end{aligned}
\end{equation}
\begin{equation}\label{eqn:t CkCj}
	\begin{aligned}
		{^{{C}_{\beta}}\boldsymbol{t}_{{C}_{\alpha}}} &= {^{C_\beta}_{G}\boldsymbol{R}}\left({^{G}\boldsymbol{p}_{C_\alpha}} - {^{G}\boldsymbol{p}_{C_\beta}}\right) \\ 
		&= {^{C}_{B}\boldsymbol{R}}{^{B_\beta}_{G}\boldsymbol{R}}\left({^{G}\boldsymbol{p}_{B_\alpha}} - {^{G}\boldsymbol{p}_{B_\beta}} + {_{B_\alpha}^{G}\boldsymbol{R}}{^{B}\boldsymbol{p}_{C}}\right) - {^{C}_{B}\boldsymbol{R}}{^{B}\boldsymbol{p}_{C}}
	\end{aligned}
\end{equation}
where the $\alpha$-th and $\beta$-th images represent the left and right base views respectively, whose suggested selection method can be defined as:
\begin{align}\label{eqn:baseframe sel}	
	(\alpha,\beta)=\mathop{argmax}\limits_{1 \leq \alpha,\beta \leq n} \left\{ \boldsymbol{\theta}_{\alpha,\beta} \right\} 
\end{align}
where ${\boldsymbol{\theta}}_{\alpha,\beta} = || [{\boldsymbol{p}_{C_\beta}\times}] {^{C_\beta}_{C_\alpha}\boldsymbol{R}}{\boldsymbol{p}_{C_\alpha}} ||$ represents the parallax between the $\alpha$-th and $\beta$-th images. Eq. \eqref{eqn:baseframe sel} can help select the base view with the largest parallax. 

According to \eqref{eqn:3Dpos PO}, the landmark-based visual residual can be defined as:
\begin{align}\label{eqn:repro PA}
	\boldsymbol{r}_{{C}_{i}}^{(uv)}=\boldsymbol{\widetilde{p}}_{{C}_{i}} - \boldsymbol{p}_{{C}_{i}}=\frac{{^{{C}_{i}}\boldsymbol{p}_{{f}}}}{{{e}_{3}^{T}}{^{{C}_{i}}\boldsymbol{p}_{{f}}}}-\boldsymbol{p}_{{C}_{i}}
\end{align}

Compared with MSCKF \cite{4209642}, Eq. \eqref{eqn:repro PA} is decoupled from 3D features, so it can directly construct measurement models with $\delta \boldsymbol{x}_{A}$ without null space projection:
\begin{align}\label{eqn:reproject PO}
	\boldsymbol{r}_{{C}_{i}}^{(uv)} \approx \boldsymbol{H}_{{x}_{i}}^{(uv)} \delta \boldsymbol{x}_{A} + \boldsymbol{n}_{i}^{(uv)}
\end{align} 
where $\boldsymbol{n}_{i}$ represents the corresponding measurement noise. $\boldsymbol{H}_{{x}_{i}}^{(uv)}$ is the Jacobian of landmark-based residual with respect to active system state, and can be defined as:
\begin{equation}\label{eqn:Hxi PO}
	\begin{aligned}
		\boldsymbol{H}_{{x}_{i}}^{(uv)}&=\frac{\partial\boldsymbol{r}_{{C}_{i}}^{(uv)}}{\partial\delta \boldsymbol{x}_{A}}
		=\frac{\partial\boldsymbol{r}_{{C}_{i}}^{(uv)}}{\partial{^{{C}_{i}}\boldsymbol{\widetilde{p}}_{{f}}}}
		\frac{\partial{^{{C}_{i}}\boldsymbol{\widetilde{p}}_{{f}}}}{\partial\delta \boldsymbol{x}_{A}}
	\end{aligned}
\end{equation}
\begin{equation}\label{eqn:Hxi PO 1}
	\begin{aligned}
		\frac{\partial\boldsymbol{r}_{{C}_{i}}^{(uv)}}{\partial{^{{C}_{i}}\boldsymbol{\widetilde{p}}_{{f}}}}
		&=\begin{bmatrix}
			\frac{1}{{e}_{3}^{T}{^{{C}_{i}}\boldsymbol{\widetilde{p}}_{{f}}}} & 0 & -\frac{{e}_{1}^{T}{^{{C}_{i}}\boldsymbol{\widetilde{p}}_{{f}}}}{{({e}_{3}^{T}{^{{C}_{i}}\boldsymbol{\widetilde{p}}_{{f}}})}^2} \\
			0 & \frac{1}{{e}_{3}^{T}{^{{C}_{i}}\boldsymbol{\widetilde{p}}_{{f}}}} & -\frac{{e}_{2}^{T}{^{{C}_{i}}\boldsymbol{\widetilde{p}}_{{f}}}}{{({e}_{3}^{T}{^{{C}_{i}}\boldsymbol{\widetilde{p}}_{{f}}})}^2}
		\end{bmatrix} 
	\end{aligned}
\end{equation}
where $\frac{\partial{^{{C}_{i}}\boldsymbol{\widetilde{p}}_{{f}}}}{\partial\delta \boldsymbol{x}_{A}}$ represent the Jacobian of 3D landmark projection in the $i$-th image relative to the active system state.
Since the online extrinsic calibration of camera-IMU is considered in this letter, $\frac{\partial{^{{C}_{i}}\boldsymbol{\widetilde{p}}_{{f}}}}{\partial\delta \boldsymbol{x}_{A}}$ is further represented as:
\begin{equation}\label{eqn:Hxi PO 2}
	\begin{aligned}
		\frac{\partial{^{{C}_{i}}\boldsymbol{\widetilde{p}}_{{f}}}}{\partial\delta \boldsymbol{x}_{A}} &= \frac{\partial{^{{C}_{i}}\boldsymbol{\widetilde{p}}_{{f}}}}{\partial \delta{^{G}_{B_i}\boldsymbol{T}}} + \frac{\partial{^{{C}_{i}}\boldsymbol{\widetilde{p}}_{{f}}}}{\partial \delta{^{G}_{B_\alpha}\boldsymbol{T}}} + \frac{\partial{^{{C}_{i}}\boldsymbol{\widetilde{p}}_{{f}}}}{\partial \delta{^{G}_{B_\beta}\boldsymbol{T}}} + \frac{\partial{^{{C}_{i}}\boldsymbol{\widetilde{p}}_{{f}}}}{\partial \delta{^{C}_{B}\boldsymbol{T}}} 
	\end{aligned}
\end{equation}
where ${^{G}_{B}\boldsymbol{T}}$ and ${^{C}_{B}\boldsymbol{T}}$ represent the pose transformation relationships of inertial frames relative to the global frame and camera frame respectively.

\subsection{Ray-based Visual Residual}\label{sec:Ray-based Visual Residual}
When a landmark is continuously tracked in $n$ image frames, its 3D position can be represented as intersection point for the rays along pixel coordinate direction of the covisibility frames:
\begin{equation}\label{eqn:ray and map}
	\begin{aligned}
		^{G}\boldsymbol{p}_{f} &= ^{G}\boldsymbol{p}_{C_1} + {Z_{C_1}}{\boldsymbol{d}_{C_1}} = \cdots \\ 
		&= ^{G}\boldsymbol{p}_{C_i} + {Z_{C_i}}{\boldsymbol{d}_{C_i}} (i\in1,\cdots,n)
	\end{aligned}
\end{equation}

Combining Eq. \eqref{eqn:ray and map} and \eqref{eqn:3Dpos PO}, we take the current frame as canonical view and align the rays of other covisibility frames into the current frame to obtain the ray depth estimation on canonical view:
\begin{equation}\label{eqn:depth PO CiL}
	\begin{aligned}
		{Z}_{C_\gamma} &= \sum_{1 \leq \gamma,i \leq n, \gamma \neq i} {\omega}_{{C_\gamma},{C_i}} {{Z}_{C_{\gamma}}^{({C_\gamma},{C_i})}} 
	\end{aligned}
\end{equation}
where $\left\{C_\gamma\right\}$ is the canonical view. ${{Z}_{C_{\gamma}}^{({C_\gamma},{C_i})}}$ represents the ray depth of canonical view obtained by using the two-view constraints \cite{WOS:000458768000003} related to the covisibility frames, and ${\omega}_{{C_\gamma},{C_i}}$ represents the weight coefficient obtained by using parallax ${\theta}_{C_\gamma,C_i}$ as the quality indicator, and can be respectively expressed as:
\begin{equation}\label{eqn:depth CiL1}
	\begin{aligned}
		{{Z}_{C_{\gamma}}^{({C_\gamma},{C_i})}} &= \frac{||[{^{{C}_{i}}\boldsymbol{t}_{{C}_{\gamma}}}\times]\boldsymbol{p}_{{C}_{i}}||}{||\left[\boldsymbol{p}_{C_i} \times\right]{^{C_i}_{C_\gamma}\boldsymbol{R}}{\boldsymbol{p}_{C_{\gamma}}}||}
	\end{aligned}
\end{equation} 
\begin{equation}\label{eqn:depth PO CiL1}
	\begin{aligned}
		{\omega}_{{C_\gamma},{C_i}} &= \frac{{\theta}_{C_\gamma,C_i}}{\sum_{1 \leq \gamma,i \leq n, \gamma \neq i}{{\theta}_{C_\gamma,C_i}}}
	\end{aligned}
\end{equation}

When a landmark is observed by the stereo visual system with fixed baseline, a relatively accurate ray depth estimation can be calculated based on Eq. \eqref{eqn:ray and map} and the installation relationship of stereo cameras.
Taking the left camera frame as canonical view, Eq. \eqref{eqn:ray and map} can be redefined as:
\begin{equation}\label{eqn:depth A stereo}
	\begin{aligned}
		{{Z}_{C_R}}{\boldsymbol{p}_{C_R}} &= {{Z}_{C_L}}{^{C_R}_{C_L}\boldsymbol{R}}{\boldsymbol{p}_{C_L}}+{^{C_R}\boldsymbol{t}_{C_L}}
	\end{aligned}
\end{equation}
where ${{Z}_{C_L}} = {{Z}_{C_\gamma}^{(s)}}$ represents the ray depth estimation of canonical view calculated by stereo cameras, and can be expressed as:
\begin{equation}\label{eqn:depth A stereo 1}
	\begin{aligned}
		{{Z}_{C_\gamma}^{(s)}} &= \frac{||[{^{{C}_{R}}\boldsymbol{t}_{{C}_{L}}}\times]\boldsymbol{p}_{{C}_{R}}||}{||\left[\boldsymbol{p}_{C_R} \times\right]{^{C_R}_{C_L}\boldsymbol{R}}{\boldsymbol{p}_{C_{L}}}||}
	\end{aligned}
\end{equation}
where ${^{C_R}_{C_L}\boldsymbol{R}}={^{C_R}_{B}\boldsymbol{R}}{_{C_L}^{B}\boldsymbol{R}}$, ${^{C_R}\boldsymbol{t}_{C_L}}={^{C_R}\boldsymbol{p}_{B}}-{^{C_R}_{B}\boldsymbol{R}}{_{C_L}^{B}\boldsymbol{R}}{^{C_L}\boldsymbol{p}_{B}}$ represent the pose transformation between left and right cameras.
\begin{figure}[htbp]
	\centering
	\subfigure[ ]{\centering \includegraphics [width=.95\columnwidth]{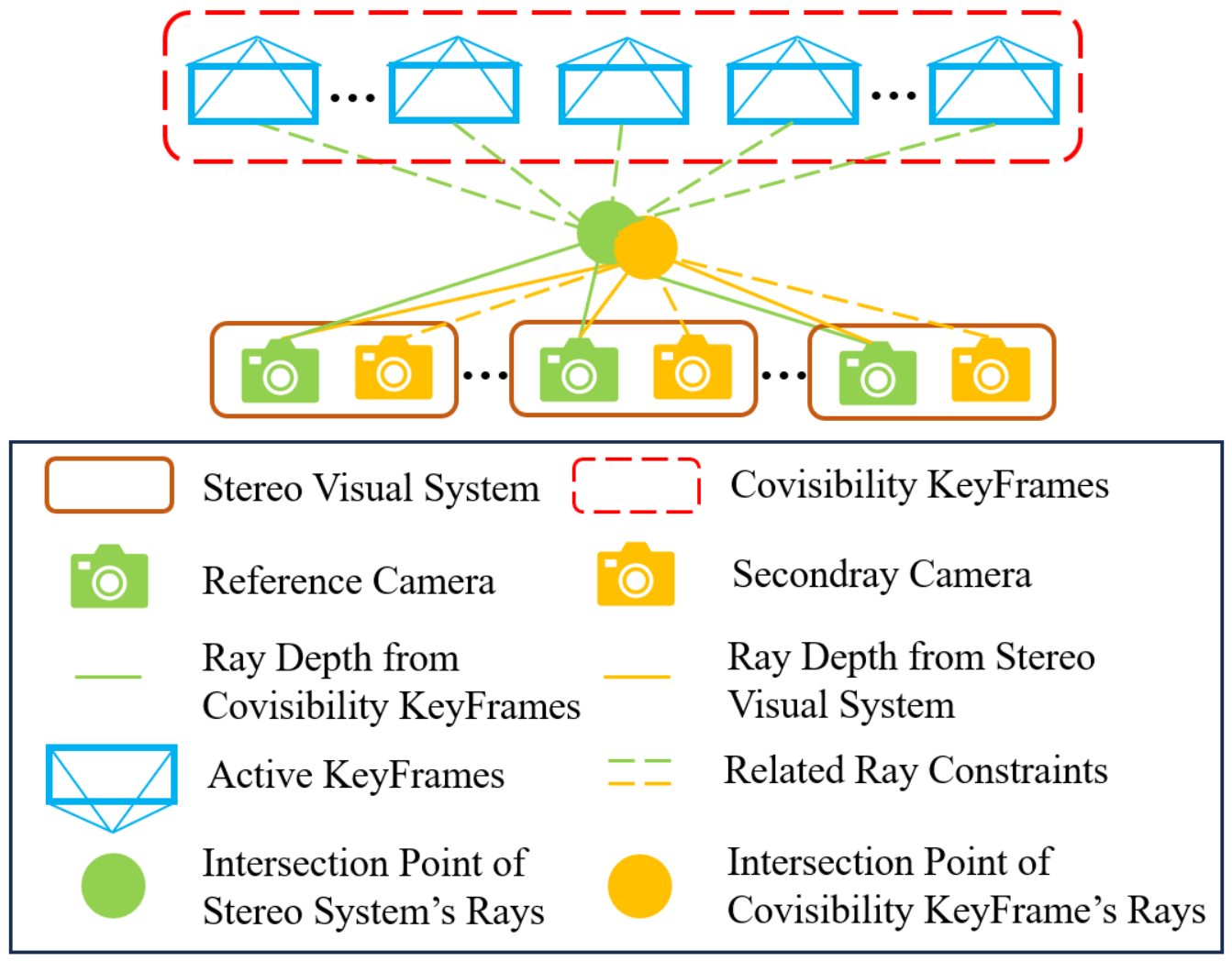}\label{fig:ray illutration}}
	
	\subfigure[ ]{\centering \includegraphics [width=.95\columnwidth]{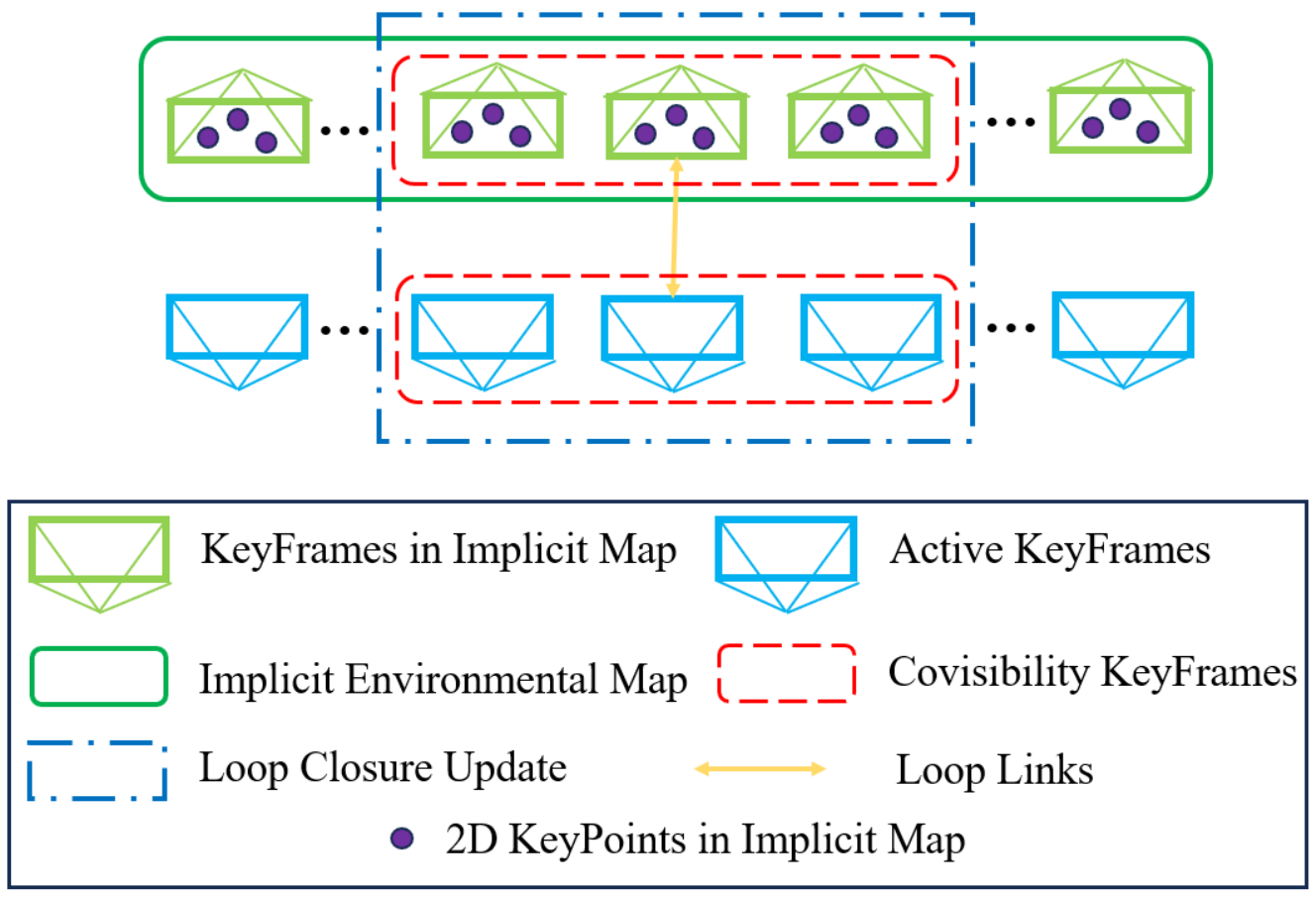}\label{fig:map illutration}}
	\caption{
		(a) Geometric representation of ray-based visual residual model (as shown in \secref{sec:Ray-based Visual Residual});
		(b) Geometric representation of implicit environmental map based loop closure (as shown in \secref{sec:Implicit Map-based Relocalization}).
	}
	\label{fig:total illutration}
\end{figure}
\begin{table*}[htbp]
	\centering
	\caption{RMSE of ATE Comparison with SOTA on EuRoC \cite{burri2016euroc} in Meters}
	\begin{threeparttable}
			\begin{tabular}{cccccccccccccc}
				\toprule
				Sequence & S/M$^{1}$ & F/O$^{1}$  & MH01 & MH02 & MH03 & MH04 & MH05 & V101 & V102 & V103 & V201 & V201 & V203 \\
				\midrule
				OKVIS$^{2}$\cite{WOS:000350472800005} & M & O & 0.16 & 0.22 & 0.24 & 0.34 & 0.47 & 0.09 & 0.20 & 0.24 & 0.13 & 0.16 & 0.29 \\
				ICE-BA$^{2,5}$\cite{liu2018ice} & S & O & 0.09 & 0.07  & 0.11 & 0.16 & 0.27 & 0.05 & 0.05 & 0.11 & 0.12 & 0.09 & 0.11 \\
				BASALT$^{2,5}$\cite{usenko2019visual} & S & O & 0.07 & 0.06 & $\boldsymbol{0.07}$ & 0.13 & \underline{0.11} & $\boldsymbol{0.04}$ & 0.05 & 0.10 & \underline{0.04} & \underline{0.05} & - \\
				VINS-Fusion$^{2,5}$\cite{qin2019general} & S & O & 0.17 & 0.15 & 0.14 & 0.27 & 0.30 & 0.08 & 0.07 & 0.10 & 0.11 & 0.09 & 0.10 \\ 
				DM-VIO$^{2}$\cite{von2022dm} & M & O & \underline{0.06} & $\boldsymbol{0.04}$ & 0.10 & \underline{0.10} & $\boldsymbol{0.10}$ & \underline{0.05} & 0.05 & \underline{0.07} & 0.03 & \underline{0.05} & 0.14 \\
				Kimera$^{3,5}$\cite{abate2023kimera2} & S & O & 0.09 & 0.11 & 0.12 & 0.15 & 0.15 & 0.06 & \underline{0.04} & 0.10 & 0.05 & 0.06 & 0.19 \\
				\midrule
				MSCKF$^{2}$\cite{4209642} & S & F & 0.42 & 0.45 & 0.23 & 0.37 & 0.48 & 0.34 & 0.20 & 0.67 & 0.10 & 0.16 & 1.13 \\
				ROVIO$^{2}$\cite{bloesch2015robust} & M & F & 0.21 & 0.25 & 0.25 & 0.49 & 0.52 & 0.10 & 0.10 & 0.14 & 0.12 & 0.14 & 0.14  \\
				OpenVINS$^{2}$\cite{9196524} & S & F & 0.07 & 0.14 & \underline{0.09} & 0.17 & 0.25 & 0.06 & 0.06 & $\boldsymbol{0.06}$ & 0.05 & \underline{0.05} & 0.13 \\
				SchurVINS$^{2}$\cite{fan2024schurvins} & S & F & $\boldsymbol{0.05}$ & 0.08 & \underline{0.09} & 0.13 & 0.13 & $\boldsymbol{0.04}$ & 0.05 & 0.08 & 0.05 & 0.08 & 0.64 \\
				Voxel-SVIO$^{4}$\cite{yuan2025voxel} & S & F & 0.09 & 0.06 & 0.10 & 0.11 & \underline{0.11} & $\boldsymbol{0.04}$ & 0.07 & $\boldsymbol{0.06}$ & 0.06 & 0.07 & 0.11 \\
				SP-VINS(O) & S & F & 0.07 & 0.06 & \underline{0.09} & 0.12 & 0.17 & 0.06 & \underline{0.04} & \underline{0.07} & 0.05 & \underline{0.05} & \underline{0.09} \\
				SP-VINS$^{5}$ & S & F & $\boldsymbol{0.05}$ & \underline{0.05} & $\boldsymbol{0.07}$ & $\boldsymbol{0.07}$ & \underline{0.11} & \underline{0.05} & $\boldsymbol{0.03}$ & $\boldsymbol{0.06}$ & $\boldsymbol{0.03}$ & $\boldsymbol{0.04}$ & $\boldsymbol{0.07}$ \\
				\bottomrule
			\end{tabular}
		\begin{tablenotes}
			\footnotesize
			\item[] $^{1}$ S/M mean stereo or monocular, F/O mean filter-based or optimization-based; $^{2}$ Results reported at \cite{fan2024schurvins};
			\item[] $^{3}$ Results reported at \cite{abate2023kimera2}; $^{4}$ Results reported at \cite{yuan2025voxel}; $^{5}$ Loop Closure is enabled.
		\end{tablenotes}
	\end{threeparttable}
	\label{tab:ATE on EuRoC in Meters}
\end{table*}
\begin{table*}[htbp]
	\centering
	\caption{RMSE of ATE Comparison with SOTA on TUM-VI \cite{schubert2018tum} in Meters}
	\begin{threeparttable}
			\begin{tabular}{cccccccccccccc}
				\toprule
				Sequence & S/M & F/O & c1 & c2 & c3 & c4 & c5 & r1 & r2 & r3 & r4 & r5 & r6 \\
				\midrule
				OKVIS$^{1}$ & M & O & 0.33 & 0.47 & 0.57 & 0.26 & 0.39 & 0.06 & 0.11 & 0.07 & 0.03 & 0.07 & \underline{0.04} \\
				BASALT$^{1,3}$ & S & O & 0.34 & 0.42 & 0.35 & 0.21 & 0.37 & 0.09 & \underline{0.07} & 0.13 & 0.05 & 0.13 & $\boldsymbol{0.02}$ \\
				VINS-Fusion$^{2,3}$ & S & O & 0.63 & 0.95 & 1.56 & 0.25 & 0.77 & 0.07 & \underline{0.07} & 0.11 & 0.04 & 0.20 & 0.08 \\
				DM-VIO$^{1}$ & M & O & \underline{0.19} & 0.47 & $\boldsymbol{0.24}$ & 0.13 & \underline{0.16} & \underline{0.03} & 0.13 & 0.09 & 0.04 & 0.06 & $\boldsymbol{0.02}$ \\ 
				\midrule
				ROVIO$^{1}$ & M & O & 0.47 & 0.75 & 0.85 & 0.13 & 2.09 & 0.16 & 0.33 & 0.15 & 0.09 & 0.12 & 0.05 \\
				OpenVINS$^{1}$ & S & F & 0.41 & 0.32 & 1.53 & 0.18 & 0.64 & 0.06 & 0.09 & 0.08 & \underline{0.03} & 0.07 & $\boldsymbol{0.02}$ \\
				SchurVINS$^{1}$ & S & F & 0.33 & 0.29 & 0.56 & 0.16 & 0.27 & 0.05 & 0.16 & 0.07 & 0.05 & \underline{0.05} & $\boldsymbol{0.02}$ \\
				Voxel-SVIO$^{2}$ & S & F & 0.29 & \underline{0.22} & \underline{0.26} & \underline{0.07} & 0.25 & 0.07 & \underline{0.07} & \underline{0.06} & 0.05 & 0.09 & \underline{0.04} \\
				SP-VINS(O) & S & F & \underline{0.19} & 0.58 & 0.34 & 0.08 & 0.24 & 0.05 & \underline{0.07} & 0.07 & \underline{0.03} & 0.07 & $\boldsymbol{0.02}$ \\
				SP-VINS$^{3}$ & S & F & $\boldsymbol{0.15}$ & $\boldsymbol{0.08}$ & 0.32 & $\boldsymbol{0.06}$ & $\boldsymbol{0.15}$ & $\boldsymbol{0.02}$ & $\boldsymbol{0.02}$ & $\boldsymbol{0.04}$ & $\boldsymbol{0.02}$ & $\boldsymbol{0.03}$ & $\boldsymbol{0.02}$ \\
				\bottomrule
			\end{tabular}
		\begin{tablenotes}
			\footnotesize
			\item[] $^{1}$ Results reported at \cite{fan2024schurvins}; $^{2}$ Results reported at \cite{yuan2025voxel}; $^{3}$ Loop closure is enabled.
		\end{tablenotes}
	\end{threeparttable}
	\label{tab:ATE Comparison SOTAs on TUM-VI}
\end{table*}

Taking the $i$-th image frame as canonical view, the corresponding ray depth estimation can be obtained respectively from stereo visual system and covisibility frames through Eq. \eqref{eqn:depth A stereo 1} and \eqref{eqn:depth PO CiL}.
The geometric representation of ray-based visual residual is shown in \figref{fig:ray illutration}, and can be constructed as follow:
\begin{equation}\label{eqn:visual redisual i}
	\begin{aligned}
		{\boldsymbol{r}_{C_i}^{(ray)}} &= {\widetilde{Z}}_{C_i} - {{Z}_{C_i}^{(s)}} \approx \boldsymbol{H}_{{x}_{i}}^{(ray)} \delta \boldsymbol{x}_{A} + \boldsymbol{n}_{i}^{(ray)}
	\end{aligned}
\end{equation}
where $\boldsymbol{H}_{{x}_{i}}^{(ray)}$ is the Jacobian of ray-based residual with respect to active system state and the derivation process is as follows:
\begin{equation}\label{eqn:jacobian depth PO}
	\begin{aligned}
		\boldsymbol{H}_{{x}_{i}}^{(ray)} &=  \frac{\partial {\boldsymbol{r}_{{C}_{i}}^{(ray)}}}{\partial \delta \boldsymbol{x}} \\
		&= \frac{\partial {\boldsymbol{r}_{{C}_{i}}^{(ray)}}}{\partial \delta{^{G}_{B_i}\boldsymbol{T}}} + \frac{\partial {\boldsymbol{r}_{{C}_{i}}^{(ray)}}}{\partial \delta{^{G}_{B_k}\boldsymbol{T}}} + \frac{\partial {\boldsymbol{r}_{{C}_{i}}^{(ray)}}}{\partial \delta{^{C}_{B}\boldsymbol{T}}}
	\end{aligned}
\end{equation}
where ${^{G}_{B_i}\boldsymbol{T}}$ and ${^{G}_{B_k}\boldsymbol{T}}$ respectively represent the poses of canonical view and other co-observable frames, $1 \leq i,k \leq n, i \neq k $.

\subsection{Implicit Map-based Loop Closure}\label{sec:Implicit Map-based Relocalization}
Since the proposed method no longer constructs 3D map points, to eliminate accumulated drift, a novel loop closure module based on implicit environmental map that seamlessly integrates with the stereo VIO system is proposed.

After each VIO update is completed, if the current frame meets following conditions: (1) the average parallax relative to latest keyframe reaches the threshold; (2) the number of tracked features drops below the threshold; (3) the average pose transformation relative to local covisibility frames exceeds the threshold, then the current frame will be added to the keyframe database in the implicit environmental map.

When establishing new keyframe, we first adopt DBoW2 \cite{galvez2012bags} for loop detection.
DBoW2 returns the loop candidates that have passed temporal and geometrical consistency check, and then adopts two-step geometric outlier rejection including 2D-2D and 3D-2D RANSAC \cite{lepetit2009ep} to confirm loop frame.

By utilizing the loop link, we can select high-quality covisibility frames of current frame from historical keyframes, and construct the implicit map-based visual residual for loop correction as:
\begin{equation}\label{eqn:reproject PO map}
	\begin{aligned}
		\boldsymbol{r}_{{C}_{i}}^{(map)} &= \boldsymbol{\widetilde{p}}_{{C}_{i}} - \boldsymbol{p}_{{C}_{i}} \approx \boldsymbol{H}_{{x}_{i}}^{(map)} \delta \boldsymbol{x} + \boldsymbol{n}_{i}^{(map)}
	\end{aligned}
\end{equation}      
where $\boldsymbol{H}_{{x}_{i}}^{(map)}$ is the Jacobian of reprojection-based residual with respect to the full system state include $\boldsymbol{x}_{A}$ and $\boldsymbol{x}_{K}$, and can be defined as:
\begin{equation}\label{eqn:Hxi PO map}
	\begin{aligned}
		\boldsymbol{H}_{{x}_{i}}^{(map)}&=\frac{\partial\boldsymbol{r}_{{C}_{i}}^{(map)}}{\partial\delta \boldsymbol{x}}
		=\frac{\partial\boldsymbol{r}_{{C}_{i}}^{(map)}}{\partial{^{{C}_{i}}\boldsymbol{\widetilde{p}}_{{f}}}}
		\frac{\partial{^{{C}_{i}}\boldsymbol{\widetilde{p}}_{{f}}}}{\partial\delta \boldsymbol{x}} \\
		\frac{\partial{^{{C}_{i}}\boldsymbol{\widetilde{p}}_{{f}}}}{\partial\delta \boldsymbol{x}} &= \frac{\partial{^{{C}_{i}}\boldsymbol{\widetilde{p}}_{{f}}}}{\partial \delta{^{G}_{B_i}\boldsymbol{T}}} + \frac{\partial{^{{C}_{i}}\boldsymbol{\widetilde{p}}_{{f}}}}{\partial \delta{^{G}_{B_\alpha}\boldsymbol{T}}} + \frac{\partial{^{{C}_{i}}\boldsymbol{\widetilde{p}}_{{f}}}}{\partial \delta{^{G}_{B_\beta}\boldsymbol{T}}} + \frac{\partial{^{{C}_{i}}\boldsymbol{\widetilde{p}}_{{f}}}}{\partial \delta{^{C}_{B}\boldsymbol{T}}}
	\end{aligned}
\end{equation}
where $\frac{\partial\boldsymbol{r}_{{C}_{i}}^{(map)}}{\partial{^{{C}_{i}}\boldsymbol{\widetilde{p}}_{{f}}}}$ is similar to \eqref{eqn:Hxi PO 1}.

As shown in \figref{fig:map illutration}, we can achieve accumulate drift correction by using the implicit environmental map after successful loop detection.

\section{Benchmark Experiments}
We evaluate our method on the EuRoC \cite{burri2016euroc}, TUM-VI \cite{schubert2018tum} and KAIST \cite{jeong2019complex}, covering flying drones, handheld devices, and autonomous driving respectively.
Utilizing the EVO \cite{WOS:000458872706092}, we adopt the Absolute Translation Error (ATE) and the Relative Pose Error (RPE) of different segment length as evaluation metrics.
All experiments are conducted on a standard laptop (Intel Core i7-10875H CPU @ 2.80GHz).
\subsection{Accuracy Evaluation with SOTA Methods}
Since SP-VINS is classified as filter-based method, we first selected three SOTA visual-inertial navigation systems, i.e., OpenVINS \cite{9196524}, SchurVINS \cite{fan2024schurvins}, and Voxel-SVIO \cite{yuan2025voxel}, for comparison.
Secondly, we have supplemented some SOTA optimization-based systems according to the benchmark experiments reported at \cite{fan2024schurvins} and \cite{yuan2025voxel}. 
For a fair comparison, the results of above systems are all derived from relevant papers or manually evaluated based on the source code and configuration provided by authors.
Additionally, we conducted the ablation study, SP-VINS (O) denotes the odometry system without enabling loop closure module, while SP-VINS denotes the full system for long-term drift correction utilizing implicit environmental map.
\begin{table}[htbp]
	\centering
	\caption{RMSE of ATE on KAIST-Urban \cite{jeong2019complex} in Meters}
	\begin{threeparttable}
		\resizebox{0.49\textwidth}{!}{
			\begin{tabular}{cccccccc}
				\toprule
				Sequence & S/M & F/O & 28 & 29 & 32 & 38 & 39  \\
				\midrule
				VINS-Fusion$^{1,2}$ & S & O & 19.29 & 24.86 & - & 25.28 & 15.70 \\
				OpenVINS$^{1}$ & S & F & 11.89 & 11.23 & 14.79 & 11.30 & 12.70\\
				Voxel-SVIO$^{1}$ & S & F & 10.75 & \underline{6.84} & 9.12 & 10.39 & \underline{9.87} \\
				SP-VINS(O) & S & F & \underline{8.39} & \underline{7.42} & 7.82 & \underline{9.89} & 10.17 \\
				SP-VINS$^{2}$ & S & F & $\boldsymbol{6.52}$ & $\boldsymbol{5.13}$ & $\boldsymbol{5.27}$ & $\boldsymbol{7.46}$ & $\boldsymbol{8.81}$ \\
				\bottomrule
			\end{tabular}
		}
		\begin{tablenotes}
			\footnotesize
			\item[] $^{1}$ Results reported at \cite{yuan2025voxel}; $^{2}$ Loop closure is enabled.
		\end{tablenotes}
	\end{threeparttable}
	\label{tab:ATE Comparison SOTAs on Kaist}
\end{table}
\begin{figure}[htbp]
	\centering
	\includegraphics [width=3.4in]{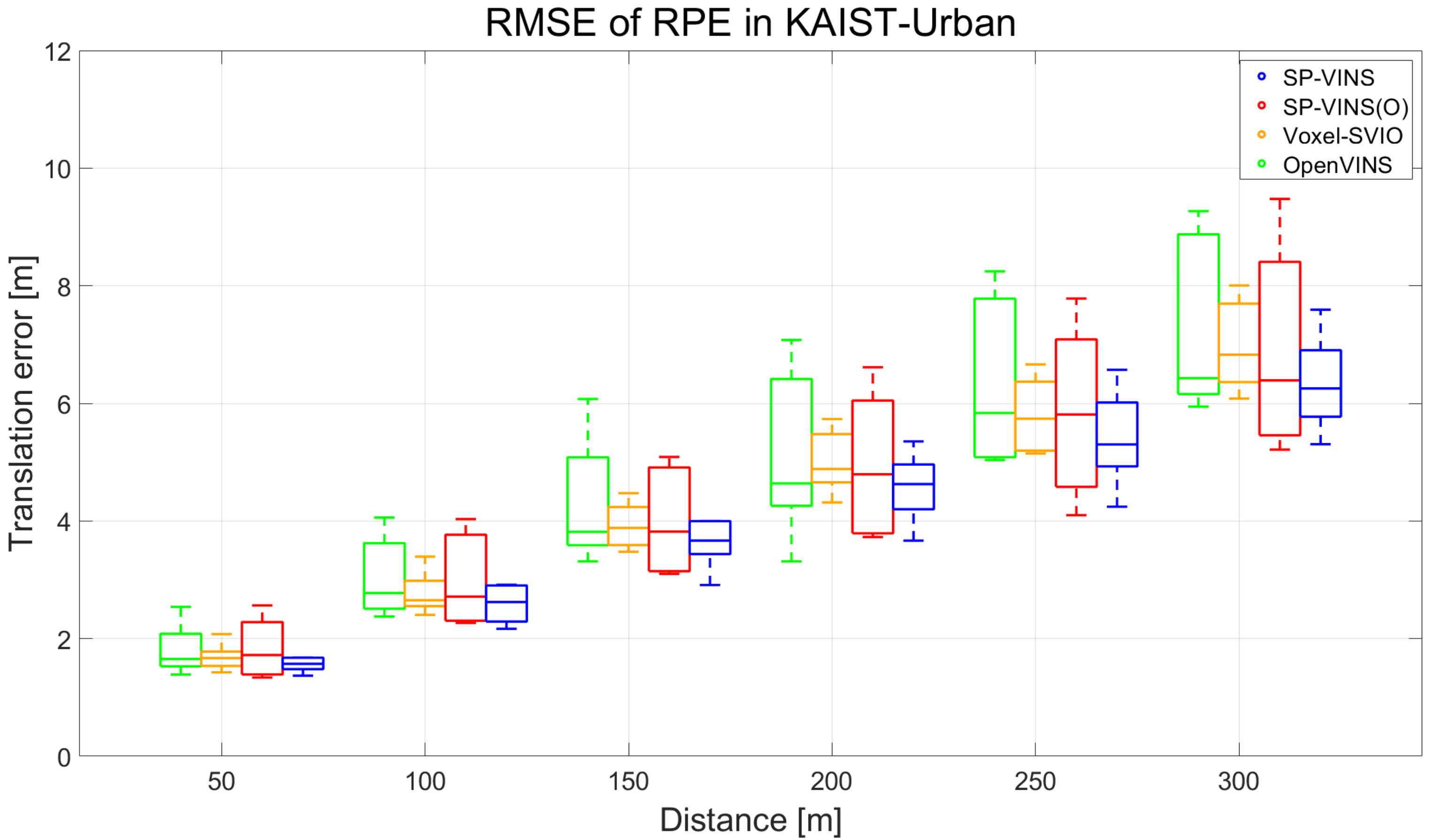} 
	\caption{
		RMSE of RPE for the comparison algorithms on KAIST-Urban.
	}
	\label{fig:RPE_kaist}
\end{figure}
\begin{figure*}[htbp]
	\centering
	\subfigure[ ]{\centering \includegraphics [width=0.31\textwidth]{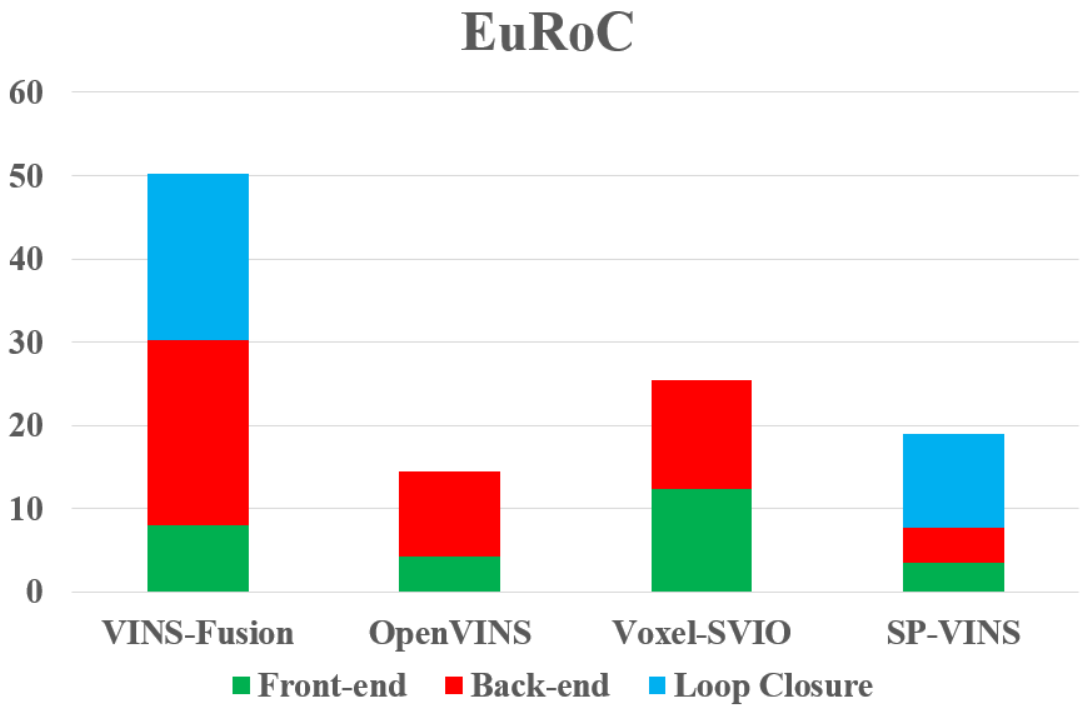}}\hfill
	\subfigure[ ]{\centering \includegraphics [width=0.31\textwidth]{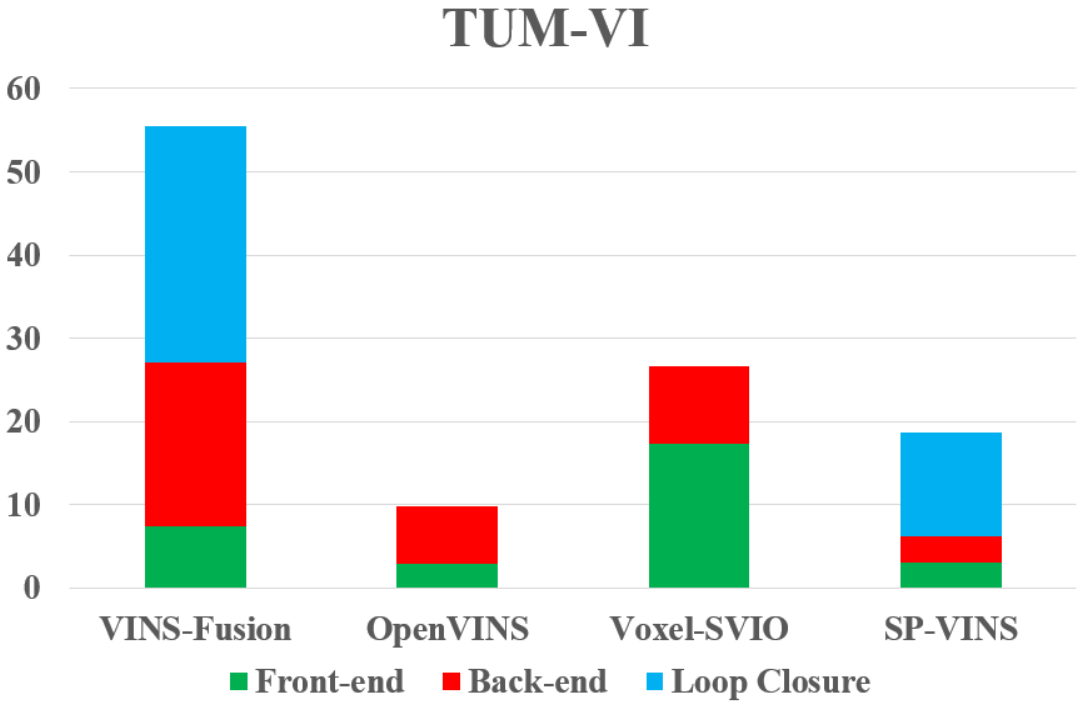}}\hfill 
	\subfigure[ ]{\centering \includegraphics [width=0.31\textwidth]{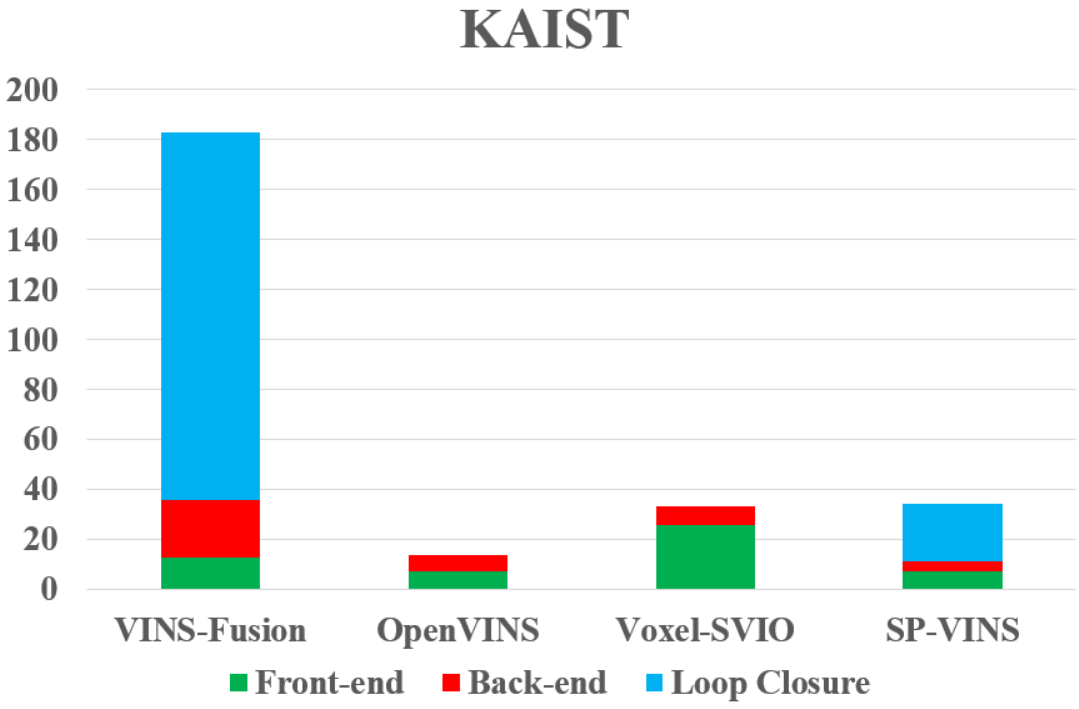}}
	\caption{
		(a)-(c) are the average runtime of comparison algorithms on different datasets (Unit: ms). Notably, the back-end runtime of OpenVINS includes enable hybrid update, while Voxel-SVIO includes the voxel-map creation and management. 
	}
	\label{fig:runtime}  
\end{figure*}
 
\tabref{tab:ATE on EuRoC in Meters}, \tabref{tab:ATE Comparison SOTAs on TUM-VI} and \tabref{tab:ATE Comparison SOTAs on Kaist} respectively demonstrates the root mean square error (RMSE) of ATE for above comparison algorithms on three datasets.
\figref{fig:RPE_kaist} shows the RMSE of RPE of different segment length for evaluated algorithms on KAIST datasets.
From \tabref{tab:ATE on EuRoC in Meters} and \tabref{tab:ATE Comparison SOTAs on TUM-VI}, SP-VINS(O) achieved accuracy close to that of SOTA filter-based systems on medium and small-scale EuRoC and TUM-VI datasets, while the full SP-VINS demonstrated the best localization performance.
As shown in \cite{yuan2025voxel}, \tabref{tab:ATE Comparison SOTAs on Kaist} and \figref{fig:RPE_kaist}, due to the accuracy damage from long-term and large-scale motion scenarios, only VINS-Fusion, OpenVINS and Voxel-SVIO were successful in the comparison systems,  while the proposed methods achieved the optimal and suboptimal localization accuracy respectively, further demonstrating robustness and advancement.

\begin{figure}[htbp]
	\centering
	\subfigure[] { \begin{minipage}{4cm} \centering \includegraphics [width=4.3cm]{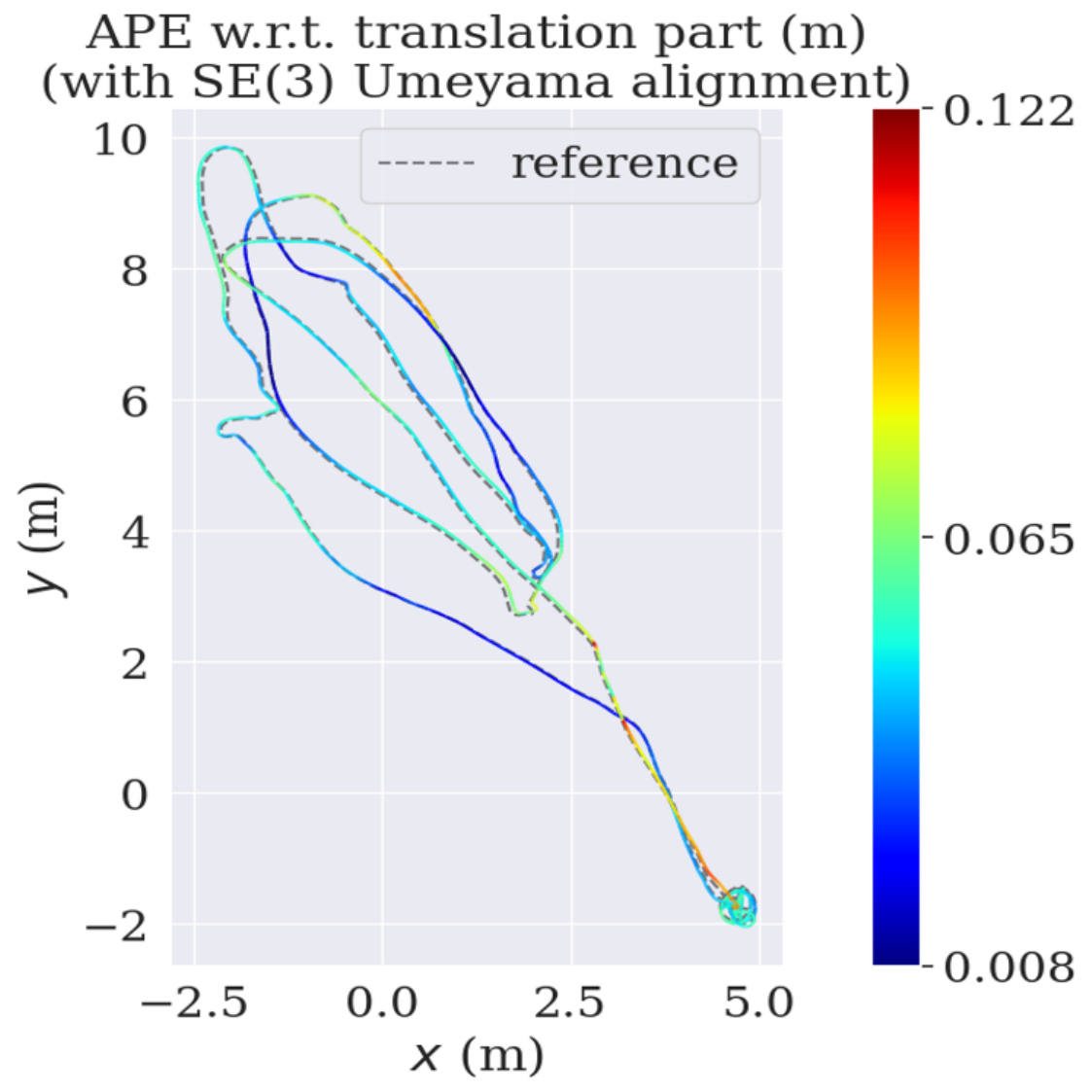}\label{fig:strong reflection} \end{minipage} } 
	\subfigure[] { \begin{minipage}{4cm} \centering \includegraphics [width=4.3cm]{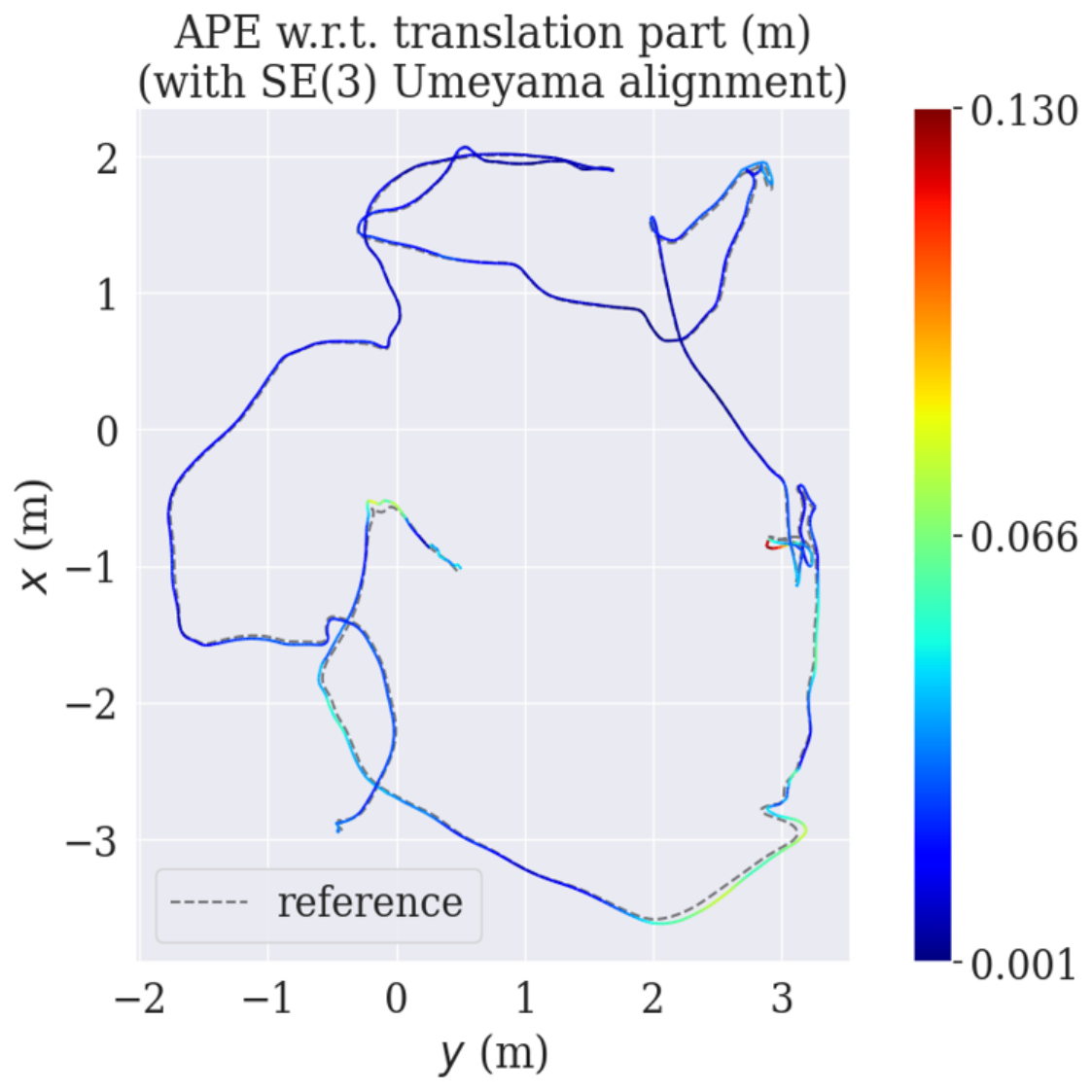}\label{fig:complex traffic} \end{minipage} } 
	\subfigure[] { \begin{minipage}{4cm} \centering \includegraphics [width=4.3cm]{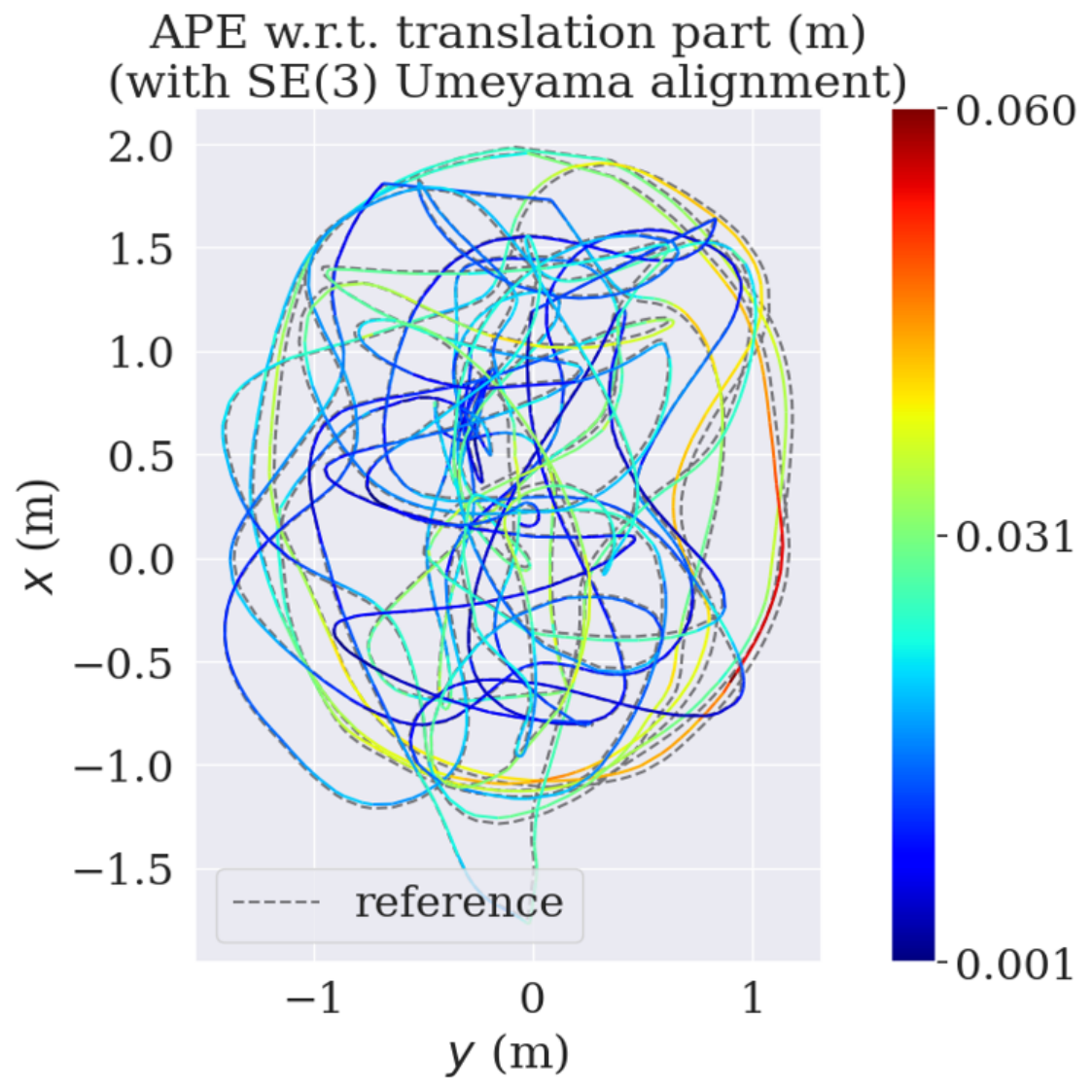}\label{fig:low texture} \end{minipage} }
	\subfigure[] { \begin{minipage}{4cm} \centering \includegraphics [width=4.3cm]{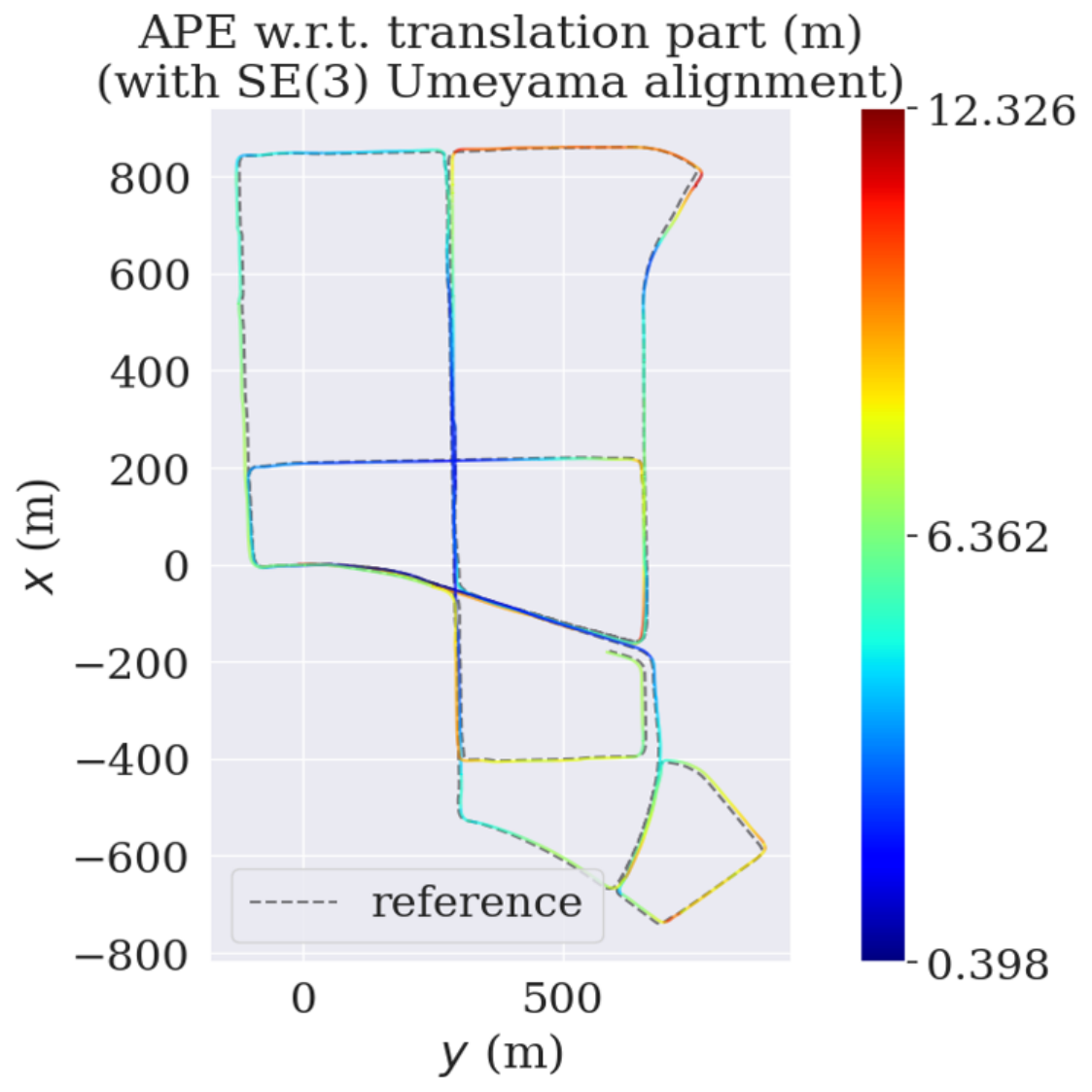}\label{fig:large depth} \end{minipage} }
	\caption{
		(a)-(d) are the comparison results between our estimated trajectories and ground truth on the exemplar sequences 'MH\_02\_easy', 'V2\_01\_medium', 'Room1\_512', 'Urban28'.
	}
	\label{fig:Challenging scenarios}
\end{figure}

Furthermore, to more intuitively demonstrate the proposed system's performance, we visualize the comparison between groud-truth and the estimated trajectories of SP-VINS.
As shown in \figref{fig:Challenging scenarios}, whether on the middle and small-scale EuRoc and TUM-VI datasets or on the large-scale KAIST dataset, the estimated trajectories of SP-VINS are highly consistent with ground-truth.   

\subsection{Efficiency Evaluation}
For the comparison algorithms that have achieved success across all datasets, we further evaluated their efficiency.
\tabref{tab:CPU usuage} and \figref{fig:runtime} respectively demonstrated the processor usage and overhead time.
\tabref{tab:CPU usuage} adopts CPU usage as evaluation metric \cite{WOS:000446394502008}, which is represented as a percentage of a single CPU core.
\figref{fig:runtime} demonstrates the time cost of following three modules: (1) Visual Front-end (image pre-process, feature extraction, matching and management, etc);(2) Back-end estimation (state propagation and augmentation, measurement update, etc); (3) Loop closure (loop detection, geometric verification, update, etc).
\begin{table}[htbp]
	\begin{center}
		\setlength{\tabcolsep}{3pt}
		\caption{Evaluation of Average CPU Usage on Different Datasets}
		\label{tab:CPU usuage}
		\begin{threeparttable}
			\resizebox{\columnwidth}{!}{
				\begin{tabular}{*{8}{c}}
					\toprule
					& \multicolumn{2}{c}{VINS-Fusion} & \multicolumn{2}{c}{OpenVINS}  & \multirow{2}{*}{Voxel-SVIO} & \multicolumn{2}{c}{SP-VINS} \\
					& ODO$^{1}$ & LC$^{2}$ & MSCKF$^{3}$ & Hybrid$^{4}$ & & ODO$^{1}$ & LC$^{2}$ \\
					\midrule
					EuRoC & 217.04 & 217.04 & 97.87 & 104.25 & 143.76 & 94.39 & 117.87 \\
					TUM-VI & 252.73 & 304.67 & 95.54 & 121.74 & 182.59 & 91.13 & 134.59 \\
					KAIST & 185.59 & 255.36 & 79.29 & 88.23 & 145.82 & 77.06 & 113.64 \\		
					\bottomrule
				\end{tabular}
			}
			\begin{tablenotes}[flushleft]
				\footnotesize
				\item[] $^{1}$ Only odometry; $^{2}$ Enable loop closure; $^{3}$ Only perform MSCKF update;
				\item[] $^{4}$ Perform hybrid update (MSCKF and SLAM).
			\end{tablenotes}
		\end{threeparttable}
	\end{center}
\end{table}

As shown in \tabref{tab:CPU usuage}, SP-VINS without the enabled LC achieves almost the lowest processor usage compared with comparison algorithms.
After enabling the LC module, the processor usage of SP-VINS remains much lower than that of VINS-Fusion equipped with LC, which can correct long-term drift while maintaining a relatively low computing consumption.

\figref{fig:runtime} further shows the specific time consumption of each module.
Benefiting from the LC module based on implicit environmental map, the computational overhead of SP-VINS is significantly reduced compared to VINS-Fusion, especially on large-scale KAIST datasets (since our method focuses only on covisibility keyframes rather than performing global pose graph optimization and bundle adjustment).

\section{Conclusion}
This letter proposes a novel autonomous navigation method, SP-VINS, which is completely decoupled from 3D map and only utilizes an implicit environment map composed of keyframes and 2D visual measurements to achieve long-term loop closure and navigation drift recovery. 
Meanwhile, this method accounts for the stereo geometric constraints and the online extrinsic calibration in degradation environment, further improving the local estimation accuracy under open-loop conditions.
We conducted a comprehensive evaluation on benchmark datasets collected from different platforms. 
The experimental results show that SP-VINS can achieve long-term and robust localization performance while maintaining a low consumption of computing resources.

\bibliographystyle{IEEEtran}

\bibliography{Ref}

\addtolength{\textheight}{-12cm}   

\end{document}